\documentclass[runningheads]{llncs}
\usepackage{graphicx}
\usepackage{amsmath,amssymb,mathrsfs} 
\usepackage{color}
\usepackage[width=122mm,left=12mm,paperwidth=146mm,height=193mm,top=12mm,paperheight=217mm]{geometry}

\usepackage[numbers, sort, comma, square]{natbib}
\usepackage{caption}
\usepackage{float, subfig}
\usepackage{tabu, booktabs, multirow}
\usepackage{changepage}
\usepackage{hyperref, url}
\usepackage{cleveref}
\usepackage{breqn}
\usepackage{afterpage}

\newcommand{\breakingcomma}{%
  \begingroup\lccode`~=`,
  \lowercase{\endgroup\expandafter\def\expandafter~\expandafter{~\penalty0 }}}

\newcolumntype{?}{!{\vrule width 1.5pt}}

\newcommand{\etal}{\textit{et al}.}

\def\Enc{\mathrm{Enc}}
\def\Dec{\mathrm{Dec}}
\def\D{\mathrm{D}}
\def\Ex{\mathbb{E}}
\def\Tr{\mathrm{Tr}}

\begin{document}
\pagestyle{headings}
\mainmatter

\title{ELEGANT: Exchanging Latent Encodings with GAN for Transferring Multiple Face Attributes}

\titlerunning{ELEGANT}
\authorrunning{Taihong Xiao\and Jiapeng Hong \and Jinwen Ma}


\author{Taihong Xiao \and Jiapeng Hong \and Jinwen Ma}
\institute{
Department of Information Science, School of Mathematical Sciences \\
and LMAM, Peking University, Beijing, 100871, China\\
\email{\{xiaotaihong, jphong\}@pku.edu.cn, \quad jwma@math.pku.edu.cn}
}

\maketitle

\begin{abstract}

Recent studies on face attribute transfer have achieved great success. A lot of models are able to transfer face attributes with an input image. However, they suffer from three limitations: (1) incapability of generating image by exemplars; (2) being unable to transfer multiple face attributes simultaneously; (3) low quality of generated images, such as low-resolution or artifacts. To address these limitations, we propose a novel model which receives two images of opposite attributes as inputs. Our model can transfer exactly the same type of attributes from one image to another by exchanging certain part of their encodings. All the attributes are encoded in a disentangled manner in the latent space, which enables us to manipulate several attributes simultaneously. Besides, our model learns the residual images so as to facilitate training on higher resolution images. With the help of multi-scale discriminators for adversarial training, it can even generate high-quality images with finer details and less artifacts. We demonstrate the effectiveness of our model on overcoming the above three limitations by comparing with other methods on the CelebA face database. A pytorch implementation is available at \url{https://github.com/Prinsphield/ELEGANT}.

\end{abstract}

\section{Introduction}

\def\picwidth{0.33}
\begin{figure}[h]
\centering

\begin{tabu}{c@{\hskip 0.05\textwidth}c}

\subfloat[removing bangs]{
\begin{minipage}[b]{\picwidth\textwidth}
\includegraphics[width=\textwidth]{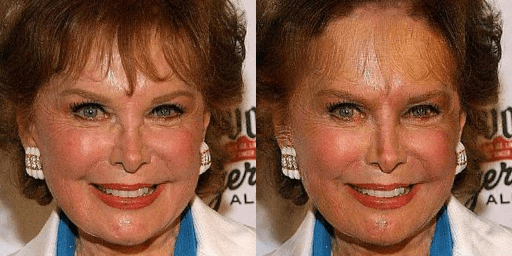}\\
\includegraphics[width=\textwidth]{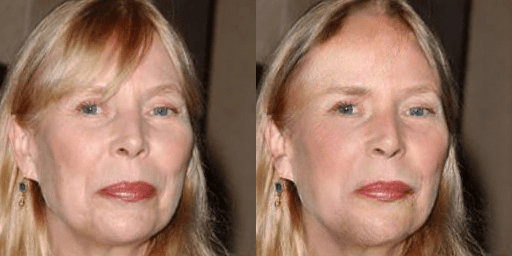}\\
\includegraphics[width=\textwidth]{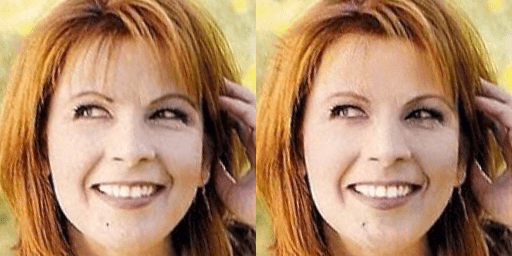}\\
\includegraphics[width=\textwidth]{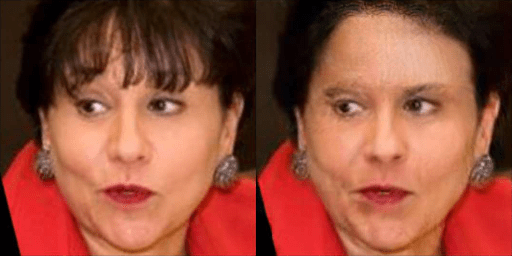}
\end{minipage}
}
&
\subfloat[adding bangs]{
\begin{minipage}[b]{\picwidth\textwidth}
\includegraphics[width=\textwidth]{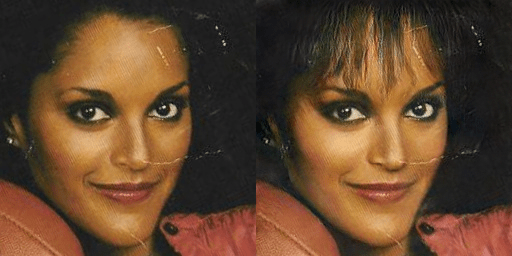}\\
\includegraphics[width=\textwidth]{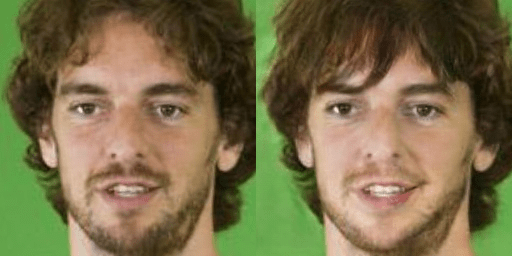}\\
\includegraphics[width=\textwidth]{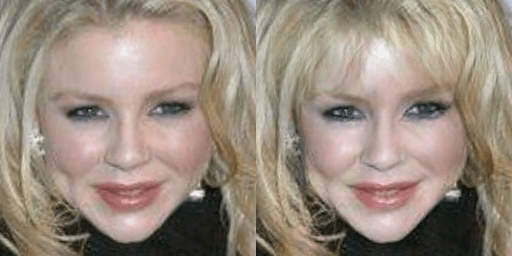}\\
\includegraphics[width=\textwidth]{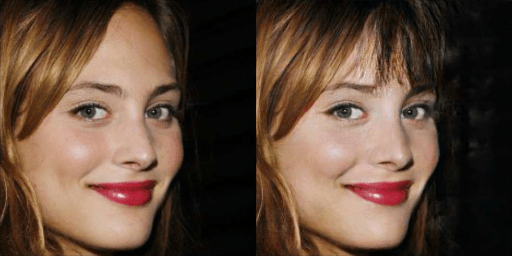}
\end{minipage}
}

\end{tabu}

\caption{Results of ELEGANT in transferring the {\tt bangs} attribute. Out of four images in a row, the bangs style of the first image is transferred to the last one.}
\label{fig:transferring_bangs}
\end{figure}

\begin{figure}[h]
\centering

\begin{tabu}{c@{\hskip 0.05\textwidth}c}

\subfloat[feminizing]{
\begin{minipage}[b]{\picwidth\textwidth}
\includegraphics[width=\textwidth]{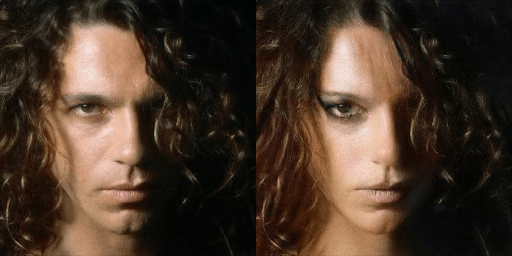}\\
\includegraphics[width=\textwidth]{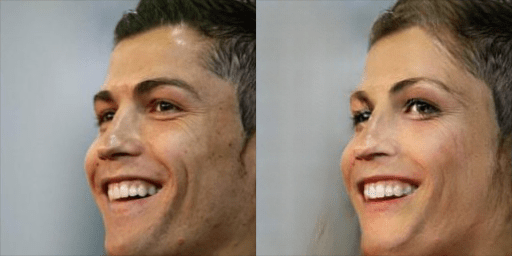}\\
\includegraphics[width=\textwidth]{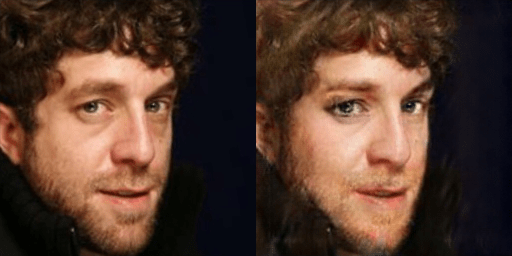}\\
\includegraphics[width=\textwidth]{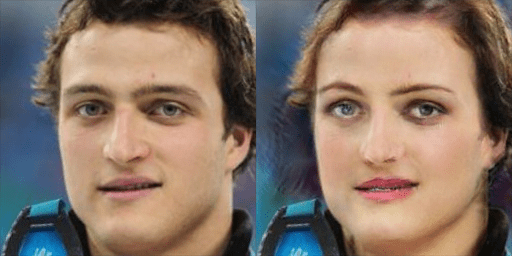}
\end{minipage}
}
&
\subfloat[virilizing]{
\begin{minipage}[b]{\picwidth\textwidth}
\includegraphics[width=\textwidth]{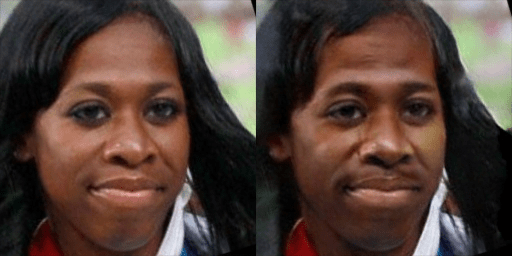}\\
\includegraphics[width=\textwidth]{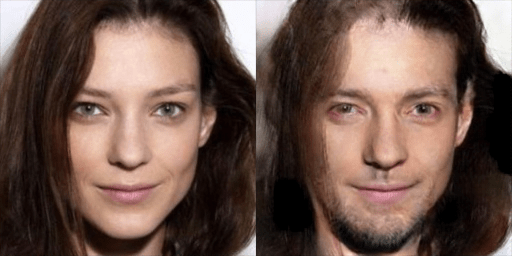}\\
\includegraphics[width=\textwidth]{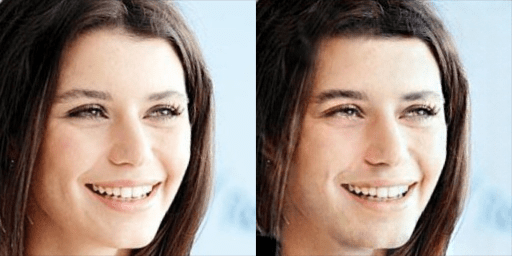}\\
\includegraphics[width=\textwidth]{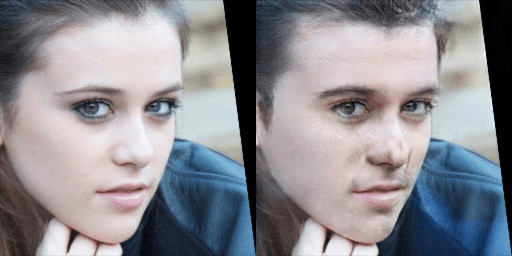}
\end{minipage}
}

\end{tabu}

\caption{Results of ELEGANT in transferring the {\tt gender} attribute.}
\label{fig:transferring_gender}
\end{figure}

\begin{figure}[h]
\centering

\begin{tabu}{c@{\hskip 0.05\textwidth}c}

\subfloat[removing eyeglasses]{
\begin{minipage}[b]{\picwidth\textwidth}
\includegraphics[width=\textwidth]{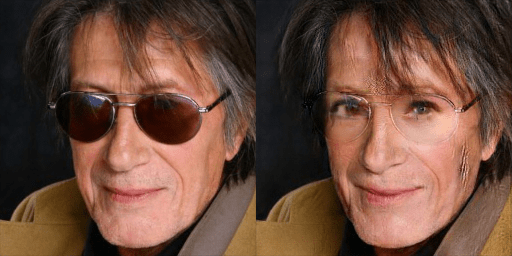}\\
\includegraphics[width=\textwidth]{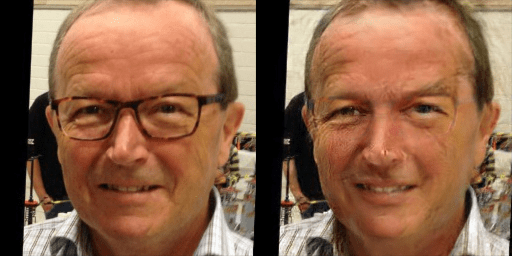}\\
\includegraphics[width=\textwidth]{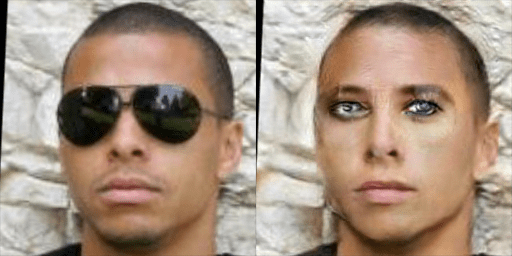}\\
\includegraphics[width=\textwidth]{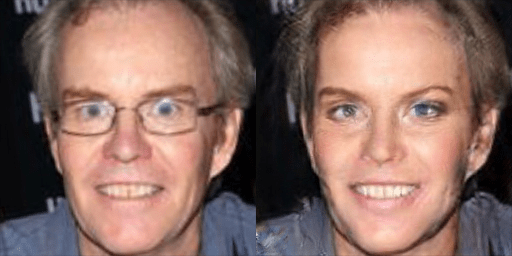}
\end{minipage}
}
&
\subfloat[adding eyeglasses]{
\begin{minipage}[b]{\picwidth\textwidth}
\includegraphics[width=\textwidth]{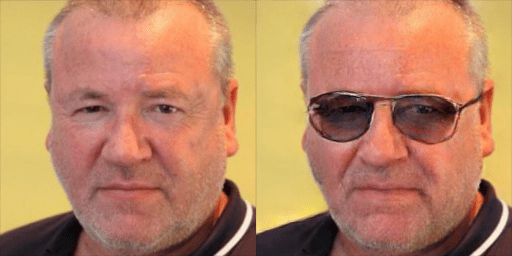}\\
\includegraphics[width=\textwidth]{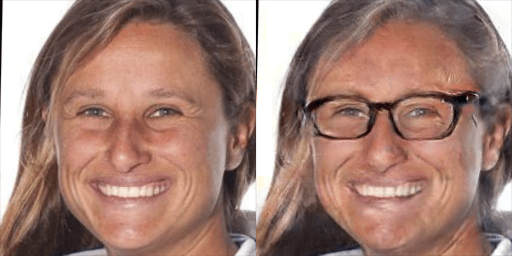}\\
\includegraphics[width=\textwidth]{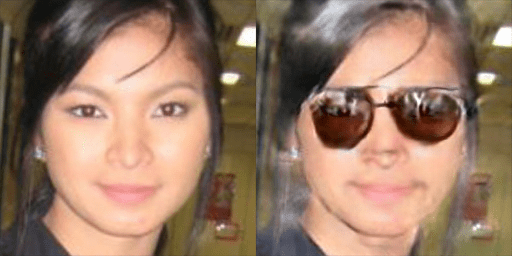}\\
\includegraphics[width=\textwidth]{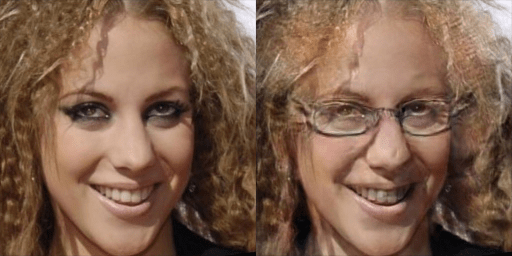}
\end{minipage}
}

\end{tabu}

\caption{Results of ELEGANT in transferring the {\tt eyeglasses} attribute. In each row, the type of eyeglasses in the first image is transferred to the last one.}
\label{fig:transferring_eyeglasses}
\end{figure}

\begin{figure}[h]
\centering

\begin{tabu}{c@{\hskip 0.05\textwidth}c}

\subfloat[removing smile]{
\begin{minipage}[b]{\picwidth\textwidth}
\includegraphics[width=\textwidth]{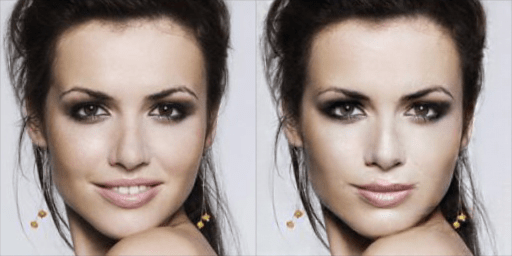}\\
\includegraphics[width=\textwidth]{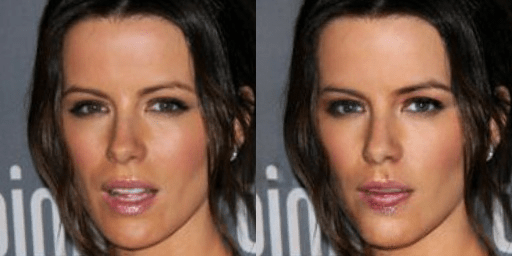}\\
\includegraphics[width=\textwidth]{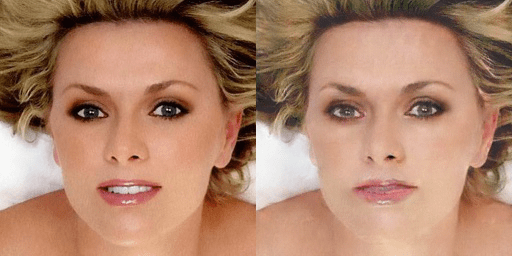}\\
\includegraphics[width=\textwidth]{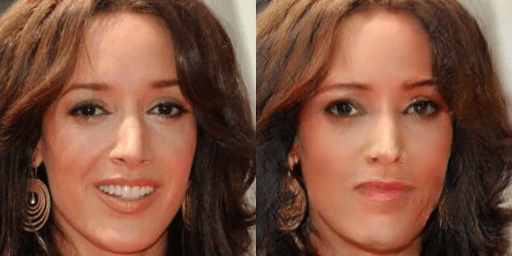}
\end{minipage}
}
&
\subfloat[adding smile]{
\begin{minipage}[b]{\picwidth\textwidth}
\includegraphics[width=\textwidth]{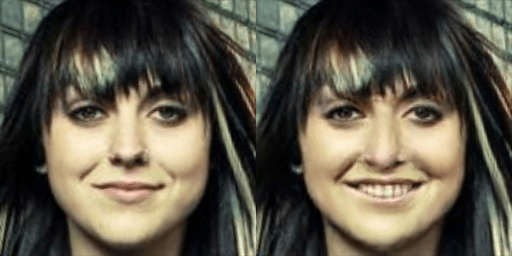}\\
\includegraphics[width=\textwidth]{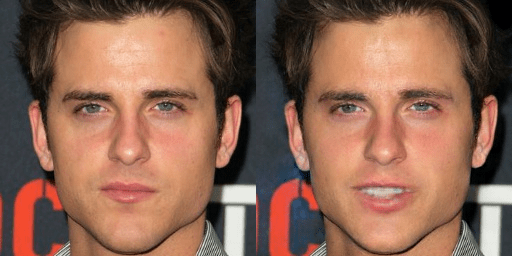}\\
\includegraphics[width=\textwidth]{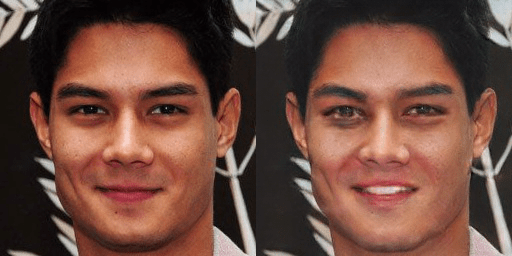}\\
\includegraphics[width=\textwidth]{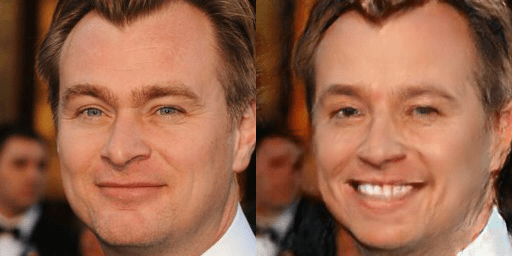}
\end{minipage}
}

\end{tabu}

\caption{Results of ELEGANT in transferring the {\tt smiling} attribute. In each row, the style of smiling of the first image is transplanted into the last one.}
\label{fig:transferring_smiling}
\end{figure}

\begin{figure}[h]
\centering

\begin{tabu}{c@{\hskip 0.05\textwidth}c}

\subfloat[black hair to non-black]{
\begin{minipage}[b]{\picwidth\textwidth}
\includegraphics[width=\textwidth]{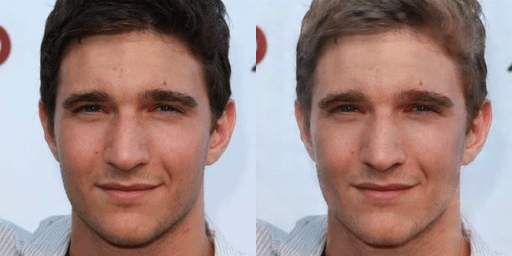}\\
\includegraphics[width=\textwidth]{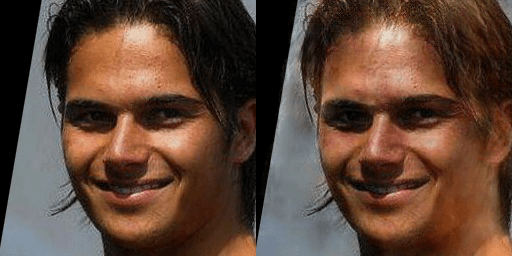}\\
\includegraphics[width=\textwidth]{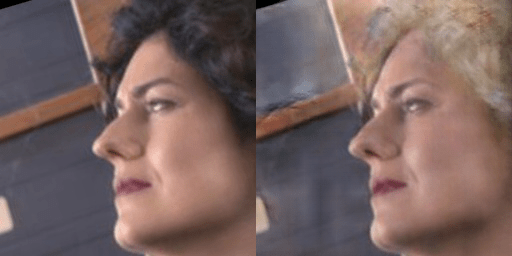}\\
\includegraphics[width=\textwidth]{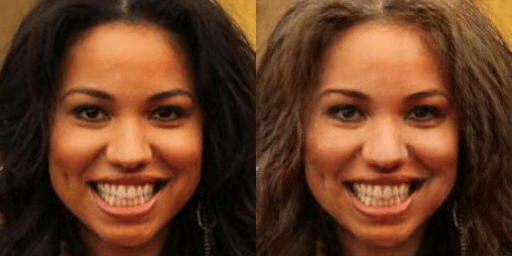}
\end{minipage}
}
&
\subfloat[non-black hair to black]{
\begin{minipage}[b]{\picwidth\textwidth}
\includegraphics[width=\textwidth]{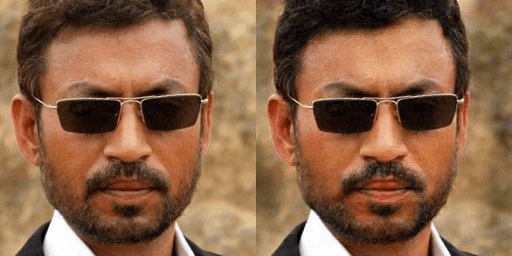}\\
\includegraphics[width=\textwidth]{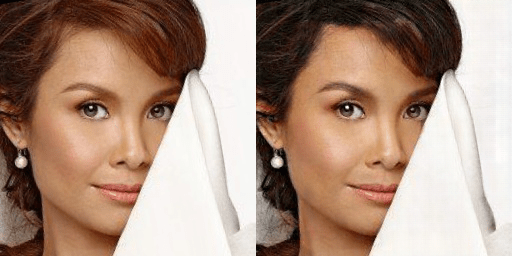}\\
\includegraphics[width=\textwidth]{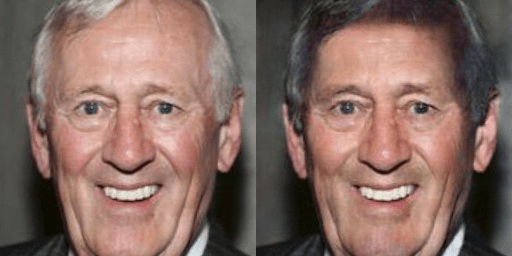}\\
\includegraphics[width=\textwidth]{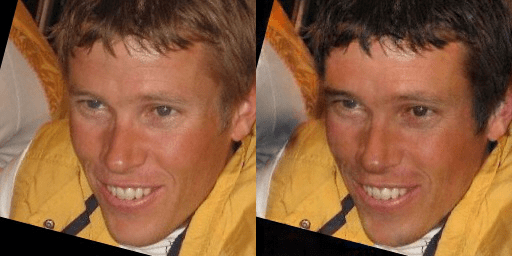}
\end{minipage}
}

\end{tabu}

\caption{Results of ELEGANT in transferring the {\tt black hair} attribute. In each row, the color of the first image turns to be the color of the third one, apart from turning the color of the third image into black.}
\label{fig:transferring_black_hair}
\end{figure}

The task of transferring face attributes is a type of conditional image generation. A source face image is modified to contain the targeted attribute, while the person identity should be preserved. As an example shown in Fig.~\ref{fig:transferring_bangs}, the bangs attribute is manipulated (added or removed) without changing the person identity. For each pair of images, the right image is purely generated from the left one, without the corresponding images in the training set.

A lot of methods have been proposed to accomplish this task, but they still suffer from different kinds of limitations.

Gardner~\etal~\cite{gardner2015deep} has proposed a method called Deep Manifold Traversal that was able to approximate the natural image manifold and compute the attribute vector from the source domain to the target domain by using maximum mean discrepancy (MMD)~\cite{gretton2012kernel}. By this method, the attribute vector is a linear combination of the feature representations of training images extracted from VGG-19~\cite{simonyan2015very} network. However, it suffers from unbearable time and memory cost, and thus is not useful in practice.

Under the Linear Feature Space assumptions~\cite{bengio2013better}, one can transfer face attribute in a much simpler manner~\cite{upchurch2016deep}: adding an attribute vector to the original image in the feature space, and then obtaining the solution in the image space inversely from the computed feature. For example, transferring a no-bangs image $B$ to a bangs image $A$ would be formulated as $A = f^{-1}(f(B) + v_{bangs})$, where $f$ is a mapping (usually deep neural networks) from the image space to the feature space, and the attribute vector $v_{bangs}$ can be computed as the difference between the cluster centers of features of bangs images and no-bangs images. The universal attribute vector is applicable to a variety of faces, leading to the same style of bangs in the generated face images. But there are many styles of bangs. Fig.~\ref{fig:transferring_bangs} would be a good illustration of different styles of bangs. Some kinds of bangs are thick enough to cover the entire forehead, some tend to go either left or right side, exposing the other half forehead, and some others may divide from the middle, etc.

To address the diversity issue, the Visual Analogy-Making~\cite{reed2015deep} has used a pair of reference images to specify the attribute vector. Such a pair of images consists of two images of the same person where one has one certain attribute and the other one does not. This method could increase the richness and diversity of generated images, however, it is usually hard to obtain a large quantity of such paired images. For example, if transferring the attribute gender over face images, we need to obtain both male and female images of a same person, which is impossible. (See Fig.~\ref{fig:transferring_gender})

Recently, more and more methods based on GANs~\cite{goodfellow2014generative} have been proposed to overcome this difficulty~\cite{perarnau2016invertible, zhu2016generative, isola2017image}. The task of face attribute transfer can be viewed as a kind of image-to-image translation problem. Images with/without one certain attribute lies in different image domains. The dual learning approaches  \cite{he2016dual, shen2017learning, kim2017learning, Yi_2017_ICCV, zhu2017unpaired} have been further exploited to map between source image domain and target image domain. The maps between the two domains are continuous and inverse to each other under the cycle consistency loss. According to the Invariance of Domain Theorem~\footnote{\url{https://en.wikipedia.org/wiki/Invariance_of_domain}}, the intrinsic dimensions of two image domains should be the same. This leads to a contradiction, because the intrinsic dimensions of two image domains are not always the same. Taking transferring eyeglasses (Fig.~\ref{fig:transferring_eyeglasses}) as an example, domain $A$ contains face images wearing eyeglasses, and domain $B$ contains face images wearing no eyeglasses. The intrinsic dimension of $A$ is larger than that of $B$ due to the variety of eyeglasses.

Some other methods~\cite{taigman2016unsupervised, liu2017unsupervised, DBLP:conf/bmvc/ZhouXYFHH17} are actually the variants of combinations of GAN and VAE. These models employ the autoencoder structure for image generation instead of using two maps interconnecting two image domains. They successfully bypass the problem of unequal intrinsic dimensions. However, most of these models are limited to manipulating only one face attribute each time.

To control multiple attributes simultaneously, lots of conditional image generation methods~\cite{perarnau2016invertible, lample2017fader, choi2018stargan, zhao2018modular} receive image labels as conditions. Admittedly, these models could transfer several attributes at the same time, but fail to generate images by exemplars, that is, generating images with exactly the same attributes in another reference image. Consequently, the style of attributes in the generated image might be similar, thus lacking of richness and diversity.

BicycleGAN~\cite{zhu2017toward} introduces a noise term to increase the diversity, but fails to generate images of specified attributes. TD-GAN~\cite{wang2017tag} and DNA-GAN~\cite{xiao2018dna} can generate images by exemplars. But TD-GAN requires explicit identity information in the label so as to preserve the person identity, which limits its application in many datasets without labeled identity information. DNA-GAN suffers from the training difficulty on high-resolution images. There also exist many other methods~\cite{li2016deep}, however, their results are not visually satisfying, either low-resolution or lots of artifacts in the generated images.

\section{Purpose and Intuition}
\label{sec:purpose_and_intuition}

As discussed above, there are many approaches to transferring face attributes. However, most of them suffer from one or more following limitations:
\begin{enumerate}
	\item Incapability of generating image by exemplars;
	\item Being unable to transfer multiple face attributes simultaneously;
	\item Low quality of generated images, such as low-resolution or artifacts.
\end{enumerate}

To overcome these three limitations, we propose a novel model integrated with different advantages for multiple face attribute transfer.

To generate images by exemplars, a model must receive a reference for conditional image generation. Most of previous methods~\cite{perarnau2016invertible, lu2017conditional, lample2017fader, choi2018stargan} use labels directly for guiding conditional image generation. But the information provided by a label is very limited, which is not commensurate with the diversity of images of that label. Various kinds of smiling face images can be classified into {\tt smiling}, but cannot be generated inversely from the same label {\tt smiling}. So we set the latent encodings of images as the reference as the encodings of an image can be viewed as a unique identifier of an image given the encoder. The encodings of reference images are added to inputs so as to guide the generation process. In this way, the generated image will have exactly the same style of attributes in the reference images.

For manipulating multiple attributes simultaneously, the latent encodings of an image can be divided  into different parts, where each part encodes information of a single attribute~\cite{xiao2018dna}. In this way, multiple attributes are encoded in a disentangled manner. When transferring several certain face attributes, the encodings parts corresponding to those attributes should be changed.

To improve the quality of generated images, we adopt the idea of residual learning~\cite{he2016deep, shen2017learning} and multi-scale discriminators~\cite{wang2018high}. The local property of face attributes is unique in the task of face attributes transfer, contrast to the task of image style transfer~\cite{gatys2015a}, where the image style is a holistic property. Such property allows us to modify only a local part of the image so as to transfer face attributes, which helps alleviate the training difficulty. The multi-scale discriminators can capture different levels of information that is useful for the generation of both holistic content and local details.


\section{Our Method}

In this section, we formally propose our method ELEGANT, the abbreviation of Exchanging Latent Encodings with GAN for Transferring multiple face attributes.

\subsection{The ELEGANT Model}
\label{sec:elegant_model}

\begin{figure}[t]
	\centering
	\includegraphics[width=\linewidth]{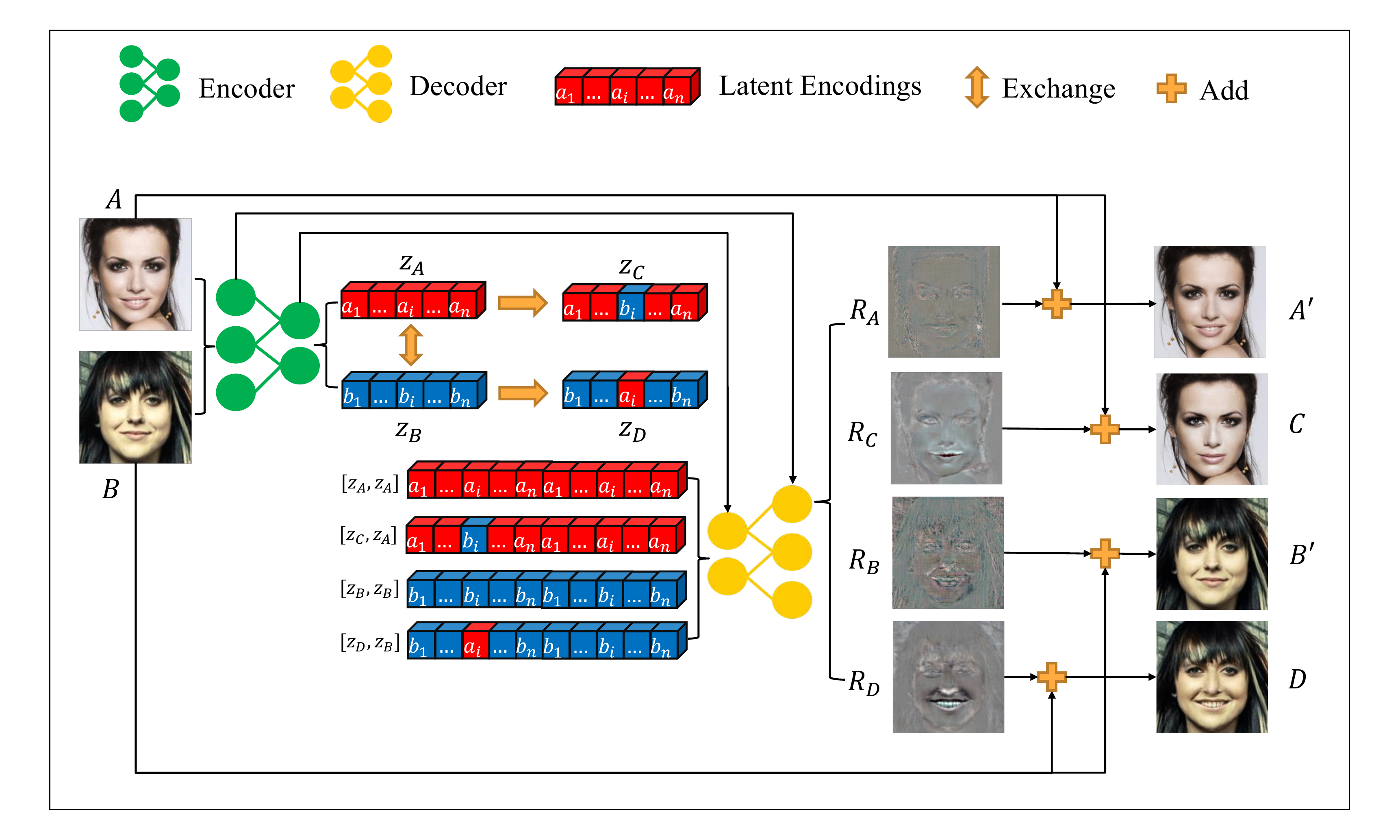}
	\caption{The ELEGANT model architecture.}
	\label{fig:model_framework}
\end{figure}

The ELEGANT model receives two sets of training images as inputs: a positive set and a negative set. In our convention, the image $A$ from the positive set has the attribute, whereas the image $B$ from the negative set does not. As shown in Fig.~\ref{fig:model_framework}, image $A$ has the attribute {\tt smiling} and image $B$ does not. The positive set and negative set need not to be paired. (The person from the positive set needs not to be the same as the one from the negative set.)

All of $n$ transferred attributes are predefined. It is not naturally guaranteed that each attribute is encoded into different parts. Such disentangled representations have to be learned. We adopt the {\it iterative training strategy}: training the model with respect to a particular attribute each time by feeding with a pair of images with opposite attribute and go over all attributes repeatedly.

When training ELEGANT about the $i$-th attribute {\tt smiling} at this iteration, a set of smiling images and another set of non-smiling images are collected as inputs. Formally, the attribute labels of $A$ and $B$ are required to be in this form $Y^A=(y_1^A, \ldots, 1_i, \ldots, y_n^A)$ and $Y^B=(y_1^B, \ldots, 0_i,\ldots,y_n^B)$, respectively.

An encoder was then used to obtain the latent encodings of images $A$ and $B$, denoted by $z_A$ and $z_B$, respectively.
\begin{align}
	z_A = \Enc(A) = [a_1, \ldots, a_i, \ldots, a_n], \qquad
	z_B = \Enc(B) = [b_1, \ldots, b_i, \ldots, b_n]
\end{align}
where $a_i$ (or $b_i$) is the feature tensor that encodes the smiling information of image $A$ (or $B$). In practice, we split the tensor $z_A$ (or $z_B$) into $n$ parts along with its channel dimension. Once obtained $z_A$ and $z_B$, we exchange the $i$-th part in their latent encodings so as to obtain novel encodings $z_C$ and $z_D$.
\begin{align}
	z_C = [a_1, \ldots, b_i, \ldots, a_n], \qquad
	z_D = [b_1, \ldots, a_i, \ldots, b_n]
\end{align}
We expect that $z_C$ is the encoding of the non-smiling version of image $A$, and $z_D$ the encodings of the smiling version of image $B$. As shown in Fig.~\ref{fig:model_framework}, $A$ and $B$ are both reference images for each other, $C$ and $D$ are generated by swapping the latent encodings.

Then we need to design a reasonable structure to decipher the latent encodings into images. As discussed in Sec.~\ref{sec:purpose_and_intuition}, it would be much better to learn the residual images rather than the original image. So we recombine the latent encodings and employ a decoder to do this job.
\begin{alignat}{8}
	\Dec([z_A, z_A]) &=& R_A, \ A' &=& A + R_A \qquad
	\Dec([z_C, z_A]) &=& R_C, \ C  &=& A + R_C \\
	\Dec([z_B, z_B]) &=& R_B, \ B' &=& B + R_B \qquad
	\Dec([z_D, z_B]) &=& R_D, \ D  &=& B + R_D
\end{alignat}
where $R_A, R_B, R_C$ and $R_D$ are residual images, $A'$ and $B'$ are reconstructed images, $C$ and $D$ are images of novel attributes, $[z_C, z_A]$ denotes the concatenation of encodings $z_C$ and $z_A$. The concatenation could be replaced by difference of two encodings, but we still use the form of concatenation, because the subtraction operation could be learnt by the $\Dec$.

Besides, we use the U-Net\cite{ronneberger2015u} structure for better visual results. The structures of $\Enc$ and $\Dec$ are symmetrical, and their intermediary layers are connected by shortcuts, as displayed in Fig.~\ref{fig:model_framework}. These shortcuts bring the original images as a context condition, so as to generate seamless novel attributes.

The $\Enc$ and $\Dec$ together act as the generator. We also need discriminators for adversarial training. However, the receptive field of a single discriminator is limited when the input image size becomes large. To address this issue, we adopt multi-scale discriminators~\cite{wang2018high}: two discriminators having identical network structures whereas operating at different image scales. We denote the discriminator operating at a larger scale by $\D_1$ and the other one by $\D_2$. $\D_1$ has a smaller receptive field compared with $\D_2$. Therefore, $\D_1$ is specialized in guiding the $\Enc$ and $\Dec$ to produce finer details, whereas $\D_2$ is adept in handling the holistic image content so as to avoid generating grimaces.

The discriminators should also receive image labels as conditional inputs. There are $n$ attributes in total. The output of discriminators in each iteration reflects how real-looking the generated images are with respect to one attribute. It is necessary to let discriminators know which attribute they are dealing with in each iteration. Mathematically, it would be a conditional form. For example, $\D_1(A|Y^A)$ represents the output score by $\D_1$ for image $A$ given its label $Y^A$. We should pay attention to the attribute labels of $C$ and $D$, since they have novel attributes.
\begin{align}
	Y^A&=(y_1^A, \ldots, 1_i, \ldots, y_n^A) \quad Y^B=(y_1^B, \ldots, 0_i,\ldots,y_n^B)\\
	Y^C&=(y_1^A, \ldots, 0_i, \ldots, y_n^A) \quad Y^D=(y_1^B, \ldots, 1_i,\ldots,y_n^B)
\end{align}
where $Y^C$ differs from $Y^A$ only in the $i$-th element, by replacing $1$ with $0$, since we do not expect $C$ to have the $i$-th attribute. The same applies to $Y^D$ and $Y^B$.

\subsection{Loss Functions}


The multi-scale discriminators $\D_1$ and $\D_2$ receive the standard adversarial loss
\begin{align}\label{eq:d_loss}
\begin{split}
	L_{\D_1} ={}& -\Ex(\log(\D_1(A|Y^A))) - \Ex(\log(1 - \D_1(C|Y^C)))\\
			    & -\Ex(\log(\D_1(B|Y^B))) - \Ex(\log(1 - \D_1(D|Y^D)))
\end{split}\\
\begin{split}
	L_{\D_2} ={}& -\Ex(\log(\D_2(A|Y^A))) - \Ex(\log(1 - \D_2(C|Y^C)))\\
			    & -\Ex(\log(\D_2(B|Y^B))) - \Ex(\log(1 - \D_2(D|Y^D)))
\end{split}\\
L_{\D} ={}& L_{\D_1} + L_{\D_2}
\end{align}
When minimizing $L_{\D}$, we are actually maximizing the scores for real images and minimizing scores for fake images in the meantime. This drives $\D_1$ and $\D_2$ to discriminate the fake images from the real ones.

As for the $\Enc$ and $\Dec$, there are two types of losses. The first type is the reconstruction loss,
\begin{equation}\label{eq:ae_reconstruction_loss}
	L_{reconstruction} = ||A-A'|| + ||B-B'||
\end{equation}
which measures how well the original input is reconstructed after a sequence of encoding and decoding. The second type is the standard adversarial loss
\begin{align}\label{eq:ae_adv_loss}
\begin{split}
	L_{adv} =& -\Ex(\log(\D_1(C|Y^C))) - \Ex(\log(\D_1(D|Y^D)))\\
			 & -\Ex(\log(\D_2(C|Y^C))) - \Ex(\log(\D_2(D|Y^D)))
\end{split}
\end{align}
which measures how realistic the generated images are. The total loss for the generator is
\begin{equation}
	L_{G} = L_{reconstruction} + L_{adv}.
\end{equation}

\section{Experiments}

\begin{figure}[t]
\centering
\includegraphics[width=0.7\textwidth]{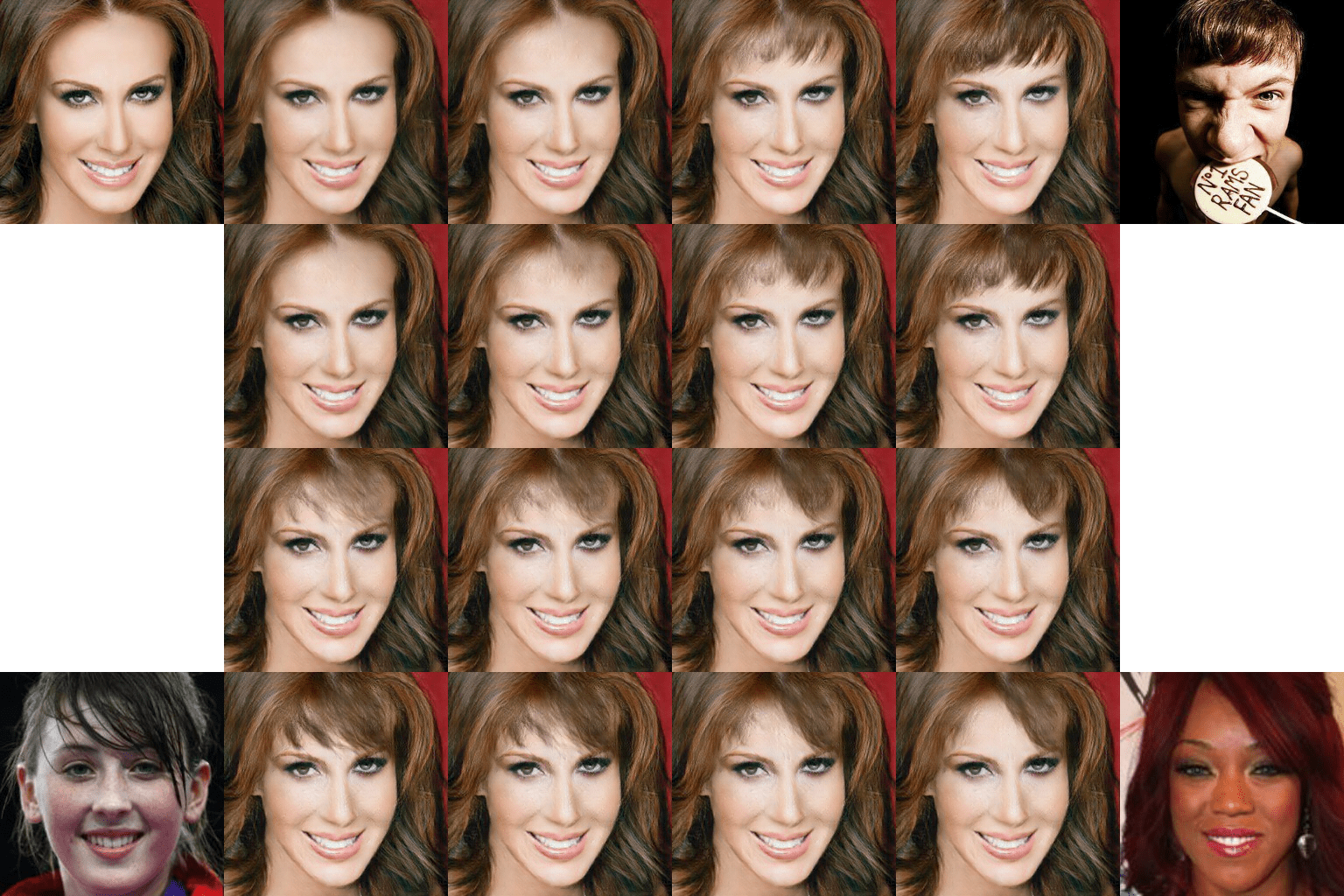}
\caption{Interpolation results of different bangs. The top-left is the original one, and those at the other three corners are reference images of different styles of bangs. The rest 16 images in the center are interpolation results.}
\label{fig:multi_bangs}
\end{figure}

In this section, we carry out different types of experiments to validate the effectiveness of our method in overcoming three limitations. First of all, we introduce the dataset and our model in details.

The CelebA~\cite{liu2015faceattributes} dataset is a large-scale face database including 202599 face images of 10177 identities, each with 40 attributes annotations and 5 landmark locations. We use the 5-point landmarks to align all face images and crop them into $256\times256$. All of the following experiments are performed at this scale.

The encoder is equipped with 5 layers of Conv-Norm-LeakyReLU block, and the decoder has 5 layers of Deconv-Norm-LeakyReLU block. The multi-scale discriminators uses 5 layers of Conv-Norm-LeakyReLU blocks followed by a fully connected layer. All networks are trained using Adam~\cite{kingma2015adam} initialized with learning rate 2e-4, $\beta_1=0.5$ and $\beta_2=0.999$. All input images are normalized into the range $[-1,1]$, and the last layer of decoder is clipped into the range $[-2,2]$ using $2\cdot\tanh$, since the maximum difference between the input image and the output image is 2. After adding the residual to the input image, we clip the output image value into $[-1,1]$ to avoid the out-of-range error.

It is worth mentioning that the Batch-Normalization (BN) layer should be avoided. ELEGANT receives two batches of images with opposite attribute as inputs, thus the moving mean and moving variance of two batches of images in each layer should make a big difference. If using BN, these running statistics in each layer will always oscillate. To overcome this issue, we replace the BN by $\ell_2$-normalization, $\hat{x} = \frac{x}{||x||_2} \cdot \alpha + \beta,$ where $\alpha$ and $\beta$ are learnable parameters. Without computing moving statistics, ELEGANT converges stably and swaps face attributes effectively.

\subsection{Face Image Generation by Exemplars}

\afterpage{\clearpage}

\begin{figure*}[!ht]
\centering
\def\picwidth{0.88}

\subfloat[bangs]{
\includegraphics[width=\picwidth\textwidth]{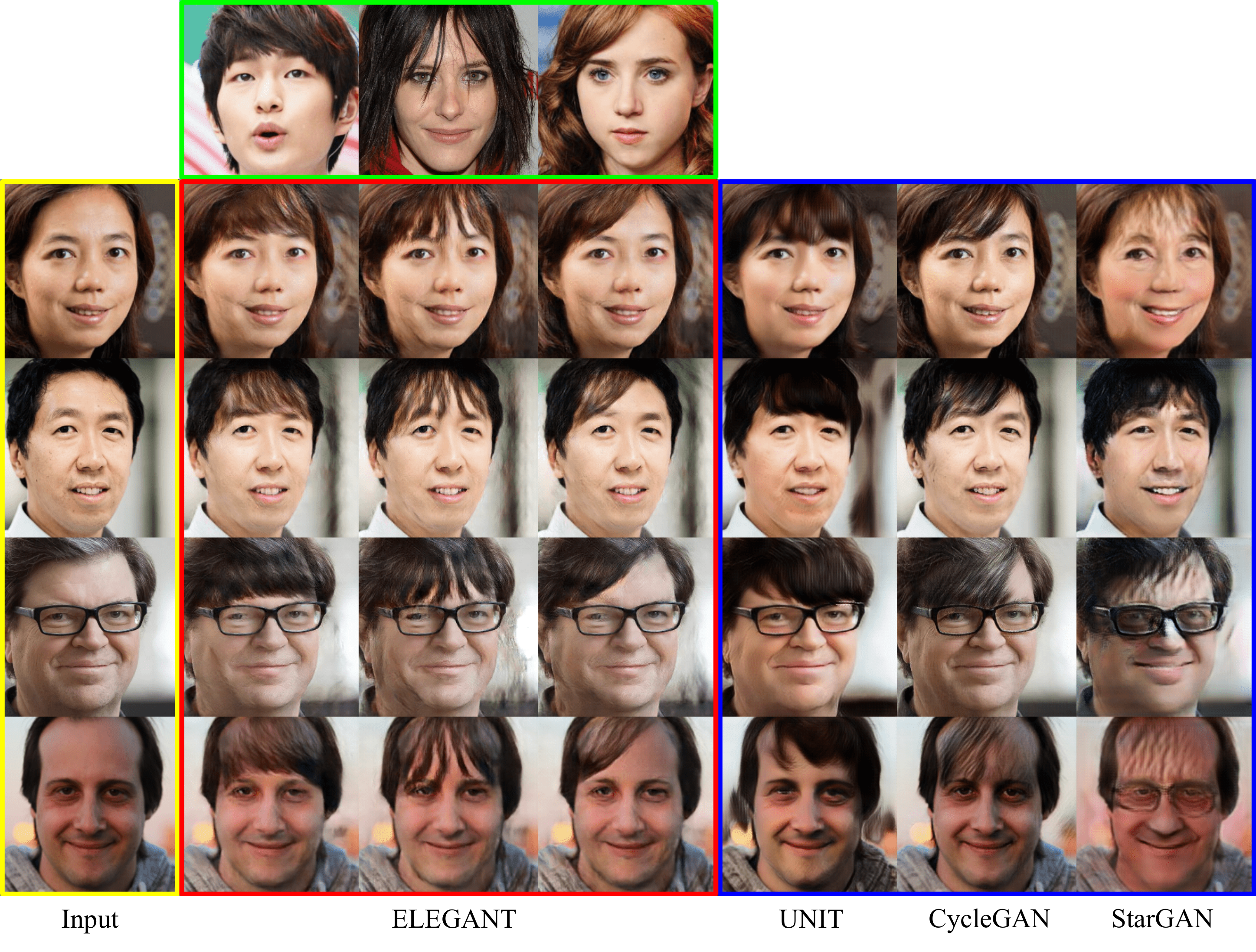}
}

\subfloat[smiling]{
\includegraphics[width=\picwidth\textwidth]{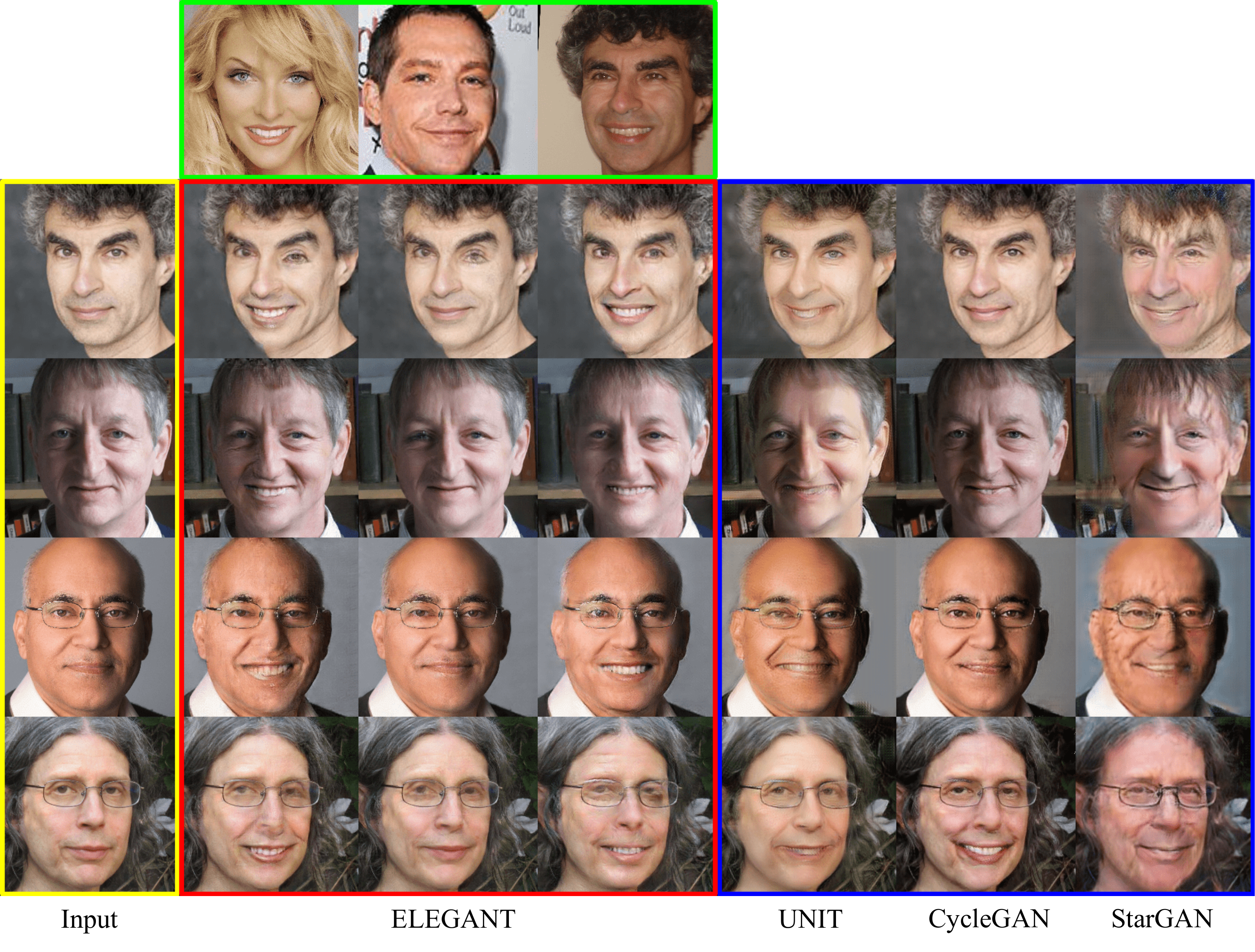}
}

\caption{Face image generation by exemplars. The yellow and green box are the input images outside the training data and the reference images, respectively. Images in the red and blue box are the results of ELEGANT and other models.}
\label{fig:exemplar}
\end{figure*}

In order to demonstrate that our model can generate face images by exemplars, we choose UNIT~\cite{liu2017unsupervised}, CycleGAN~\cite{zhu2017unpaired} and StarGAN~\cite{choi2018stargan} for comparison. As shown in Fig.~\ref{fig:exemplar}, ELEGANT can generate different face images with exactly the same style of attribute in the reference images, whereas other methods are only able to generate a common style of attribute for any input images. (The style of bangs is the same in each column in the blue box.)

An important drawback of StarGAN should be pointed out here. StarGAN could be trained to transfer multiple attributes, but when transferring only one certain attribute, it may change other attributes. For example, in the last column of Fig.~\ref{fig:exemplar}(a), Fei-Fei Li and Andrew Ng become younger when adding bangs to them. This is because StarGAN requires an unambiguous label for the input image, and these two images are both labeled as 1 in the attribute {\tt young}. However, both of them are middle-aged and cannot be simply labeled as either young or old.

The mechanism of exchanging latent encodings in the ELEGANT model effectively addresses this issue. ELEGANT focuses on the attribute that we are dealing with and does not require labels for the input images at testing phase. Moreover, ELEGANT could learn the subtle difference between different bangs style in the reference images, as displayed in Fig.~\ref{fig:multi_bangs}.

\subsection{Dealing with Multiple Attributes Simultaneously}

\begin{figure}[!htb]
\centering

\begin{tabu}{c@{\hskip 0.4\textwidth}c}
ELEGANT & DNA-GAN
\end{tabu}

\def\picwidth{0.48}
\def\skipwidth{0.03}
\subfloat[Bangs and Smiling]{
	\includegraphics[width=\picwidth\textwidth]{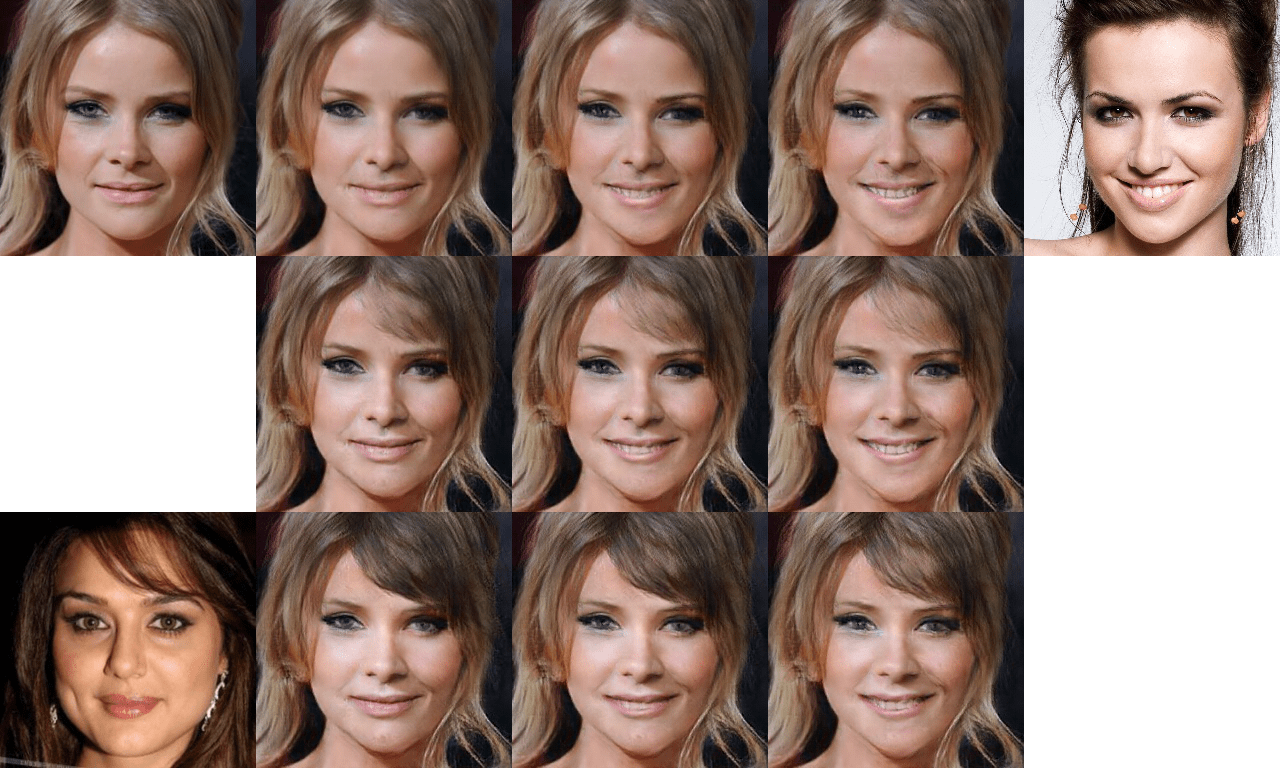} \hskip\skipwidth\textwidth
	\includegraphics[width=\picwidth\textwidth]{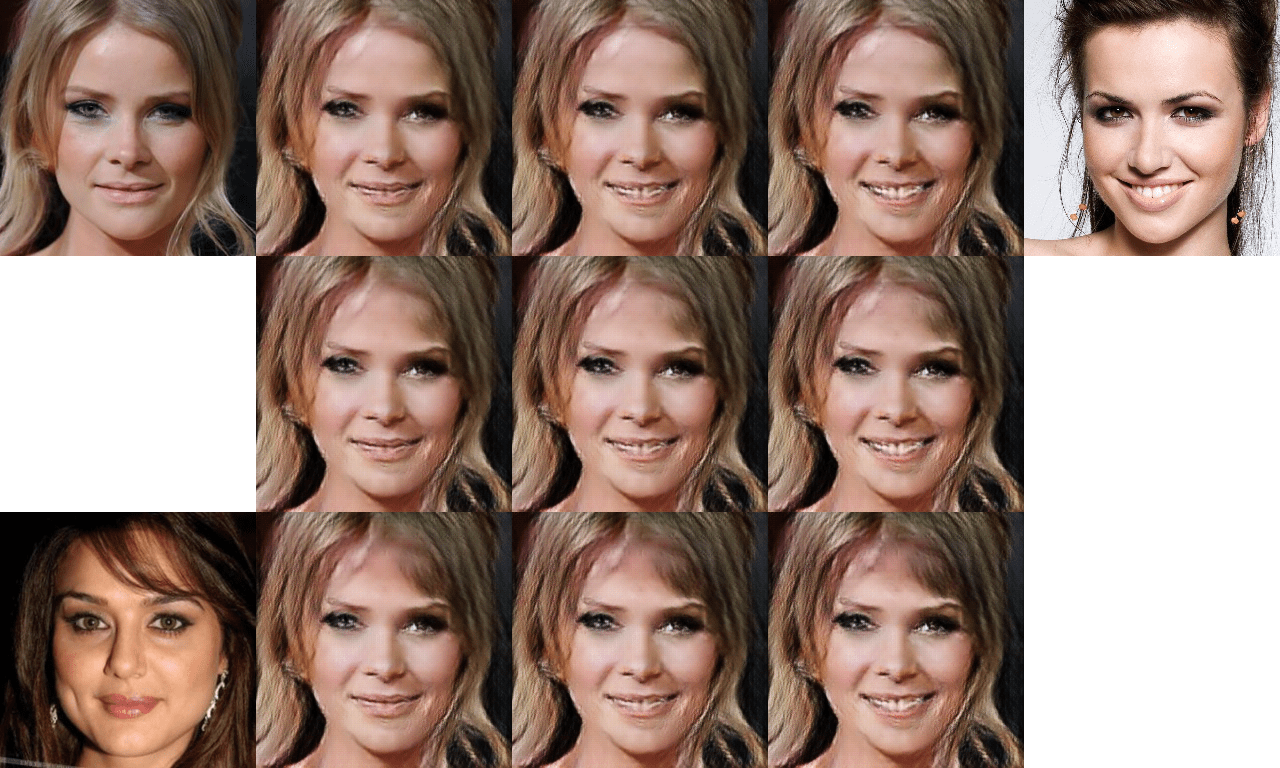}
}
\\
\subfloat[Smiling and Mustache]{
	\includegraphics[width=\picwidth\textwidth]{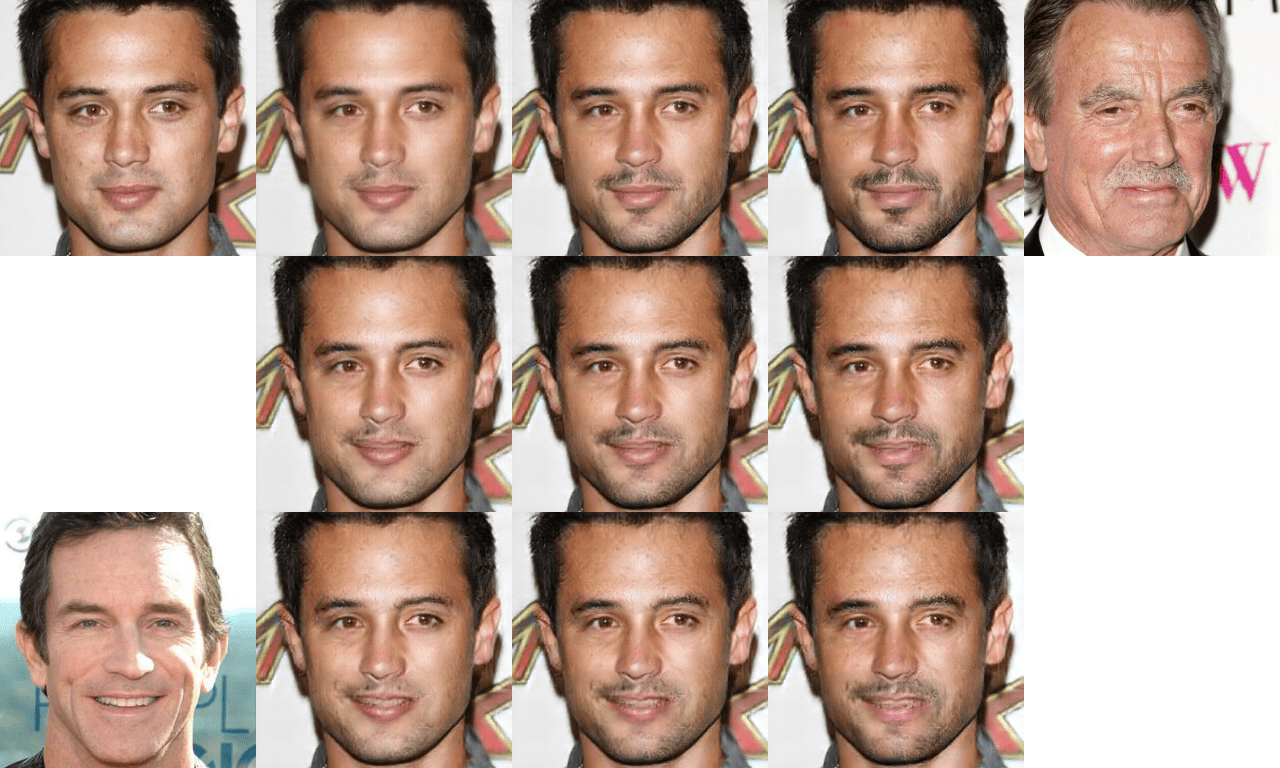} \hskip\skipwidth\textwidth
	\includegraphics[width=\picwidth\textwidth]{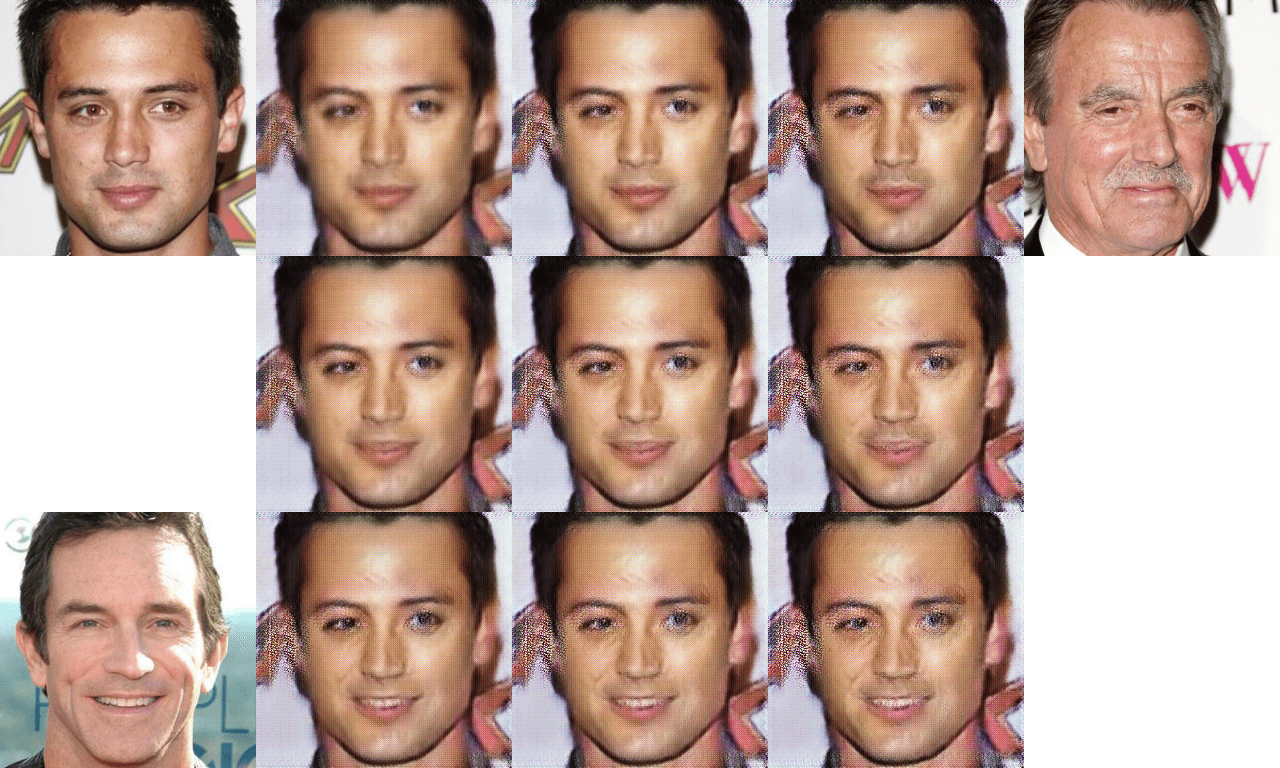}
}
\\
\subfloat[Bangs and Mustache]{
	\includegraphics[width=\picwidth\textwidth]{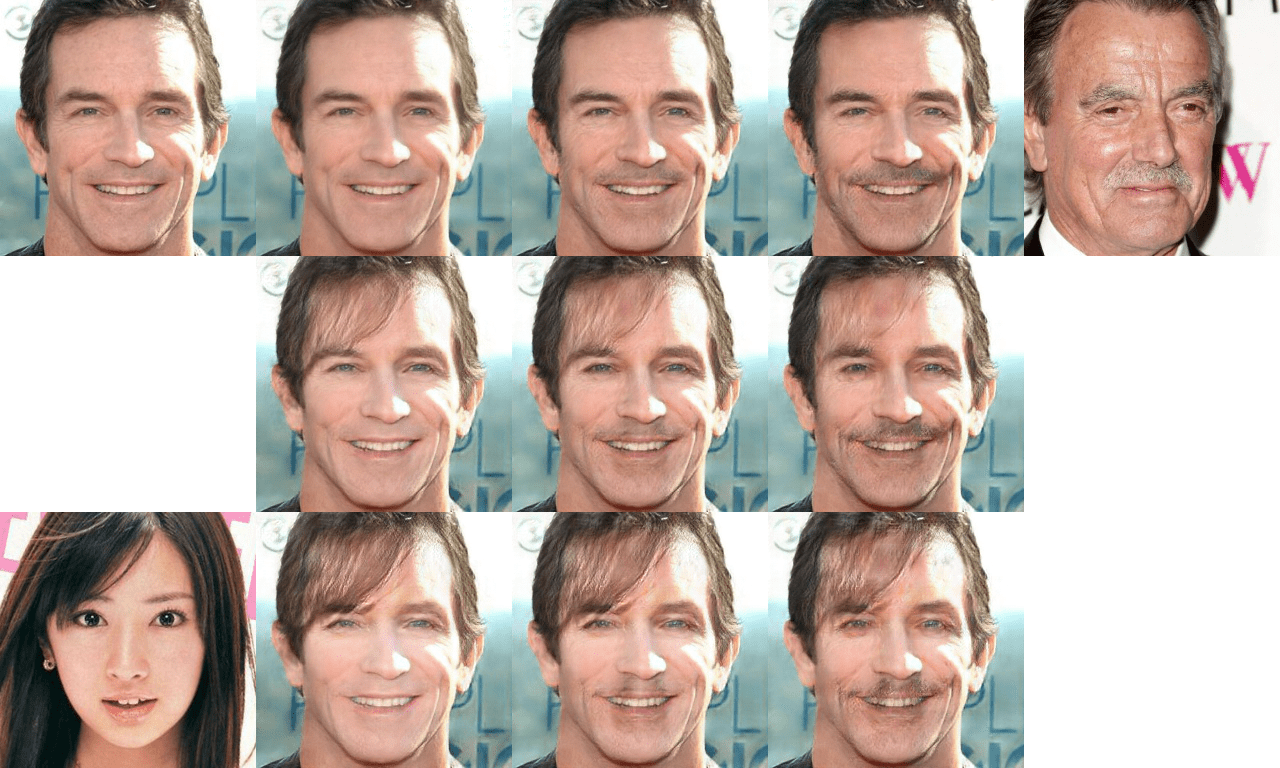} \hskip\skipwidth\textwidth
	\includegraphics[width=\picwidth\textwidth]{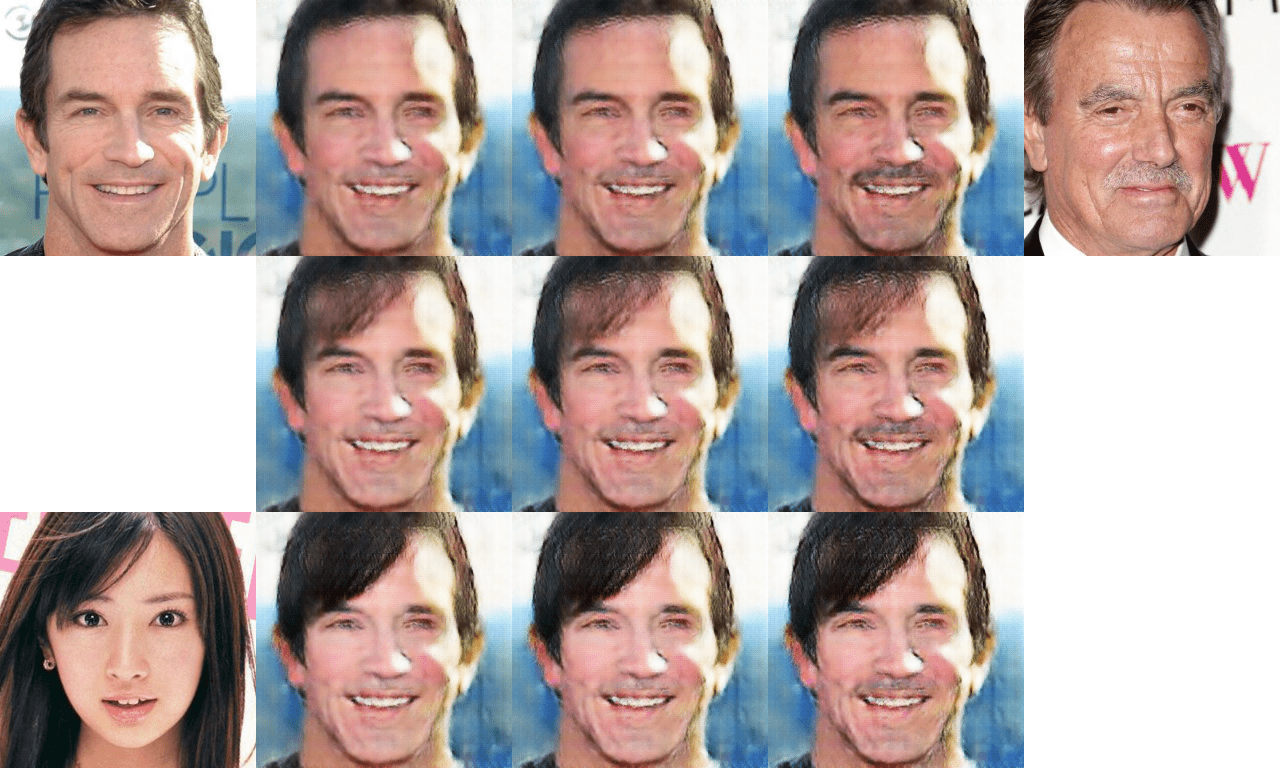}
}

\caption{Multiple Attributes Interpolation. The left and right columns are results of ELEGANT and DNA-GAN, respectively. For each picture, the top-left, bottom-left and top-right images are the original image, reference images of the first and the second attributes. The original image gradually owns two different attributes of the reference images in two directions.}
\label{fig:multi_matrix}
\end{figure}

We compare ELEGANT with DNA-GAN~\cite{xiao2018dna}, because both of them are able to manipulate multiple face attributes and generate images by exemplars. Two models are performed on the same face images and reference images with respect to three attributes. As shown in Fig.~\ref{fig:multi_matrix}, the ELEGANT is visually much better than DNA-GAN, particularly in producing finer details (zooming in for a closer look). The improvement compared with DNA-GAN is mainly the result of the residual learning and multi-scale discriminators.

Residual learning reduces training difficulty. DNA-GAN suffers from unstable training, especially on high resolution images. On one hand, this difficulty comes from an imbalance between the generator and discriminator. At the early stage of DNA-GAN training, the generator outputs nonsense so that the discriminator could easily learn how to tell generated images from real ones, which would break the balance quickly. However, ELEGANT adopts the idea of residual learning, thus the outputs of the generator are almost the same as original images at the early stage. In this way, the discriminator cannot be well trained so fast, which would help stabilize the training process. On the other hand, the burden of the generator becomes heavier than that of the discriminator as the image size goes larger. Because the output space of the generator gets larger (e.g., $256\times256\times3$), whereas the discriminator only needs to output a number as usual. However, ELEGANT effectively reduces the dimension of generator's output space by learning residual images, where a small number of pixels need to be modified.

Multi-scale discriminators improve the quality of generated images. One discriminator operating at the smaller input scale can guide the overall image content generation, and the other operating at the larger input scale can help the generator to produce finer details. (Already discussed in Sec.~\ref{sec:elegant_model})

Moreover, DNA-GAN utilizes an additional part to encode face id and background information. It is a good idea, but brings the problem of trivial solutions: two input images can be directly swapped so as to satisfy the loss constraints. Xiao~\etal~\cite{xiao2018dna} have proposed the so called annihilating operation to address this issue. But this operation leads to a distortion on the parameter spaces, which brings additional difficulty to training. ELEGANT learns the residual images that account for the changes so that the face id and background information are automatically preserved. Moreover, it removes the annihilating operation and the additional part in the latent encodings, which makes the whole framework more elegant and easy to understand.

\subsection{High-quality Generated Images}

As displayed in
Fig.~\ref{fig:transferring_bangs}~\ref{fig:transferring_gender}~\ref{fig:transferring_eyeglasses}~\ref{fig:transferring_smiling}~\ref{fig:transferring_black_hair}, we present the results of ELEGANT with respect to different attributes in a large size for a close look.
Moreover, we use the Fr\'echet Inception Distance~\cite{heusel2017gans} (FID) to measure the quality of generated images. FID measures the distance of two distributions by
\begin{equation}
	d^2 = ||\mu_1 - \mu_2||^2 + \Tr(C_1 + C_2 - 2(C_1C_2)^{1/2}).
\end{equation}
where $(\mu_1, C_1)$ and $(\mu_2, C_2)$ are means and covariance matrices of two distributions.
As shown in Table~\ref{tab:FID}, we compute the FID between the distribution of real images and generated images with respect to different attributes. ELEGANT achieves competitive results compared with other methods.

The FID score is only for reference due to two reasons.
ELEGANT and DNA-GAN can generate images by exemplars, which is much more general and difficult than other types of image translation methods. So it would be still unfair to them using any kind of qualitative measures. Besides, the reasonable qualitative measure for GAN is undetermined.

\begin{table}[t]
  \centering
  \caption{FID of Different Methods with respect to five attributes. The $+$ ($-$) represents the generated images by adding (removing) the attribute.}
    \begin{tabu}{?c?r|r|r|r|r|r|r|r|r|r?}
    \tabucline[1.5pt]{-}
    \multirow{2}[1]{*}{FID} & \multicolumn{2}{c|}{\tt bangs} & \multicolumn{2}{c|}{\tt smiling} & \multicolumn{2}{c|}{\tt mustache} & \multicolumn{2}{c|}{\tt eyeglasses} & \multicolumn{2}{c?}{\tt male} \\
    \cline{2-11}
          & \multicolumn{1}{c|}{$+$}  & \multicolumn{1}{c|}{$-$} & \multicolumn{1}{c|}{$+$}  & \multicolumn{1}{c|}{$-$}& \multicolumn{1}{c|}{$+$}  & \multicolumn{1}{c|}{$-$}& \multicolumn{1}{c|}{$+$}  & \multicolumn{1}{c|}{$-$}& \multicolumn{1}{c|}{$+$}  & \multicolumn{1}{c?}{$-$}\\
    \hline

    UNIT & 135.41 & 137.94 & 120.25 & 125.04 & 119.32 & 131.33 & 111.49 & 139.43 & 152.16 &  154.59 \\
    \hline
        CycleGAN & \bf{27.81}  & 33.22  & \bf{23.23}  & \bf{22.74}  & 43.58  & 55.49 & \bf{36.87} & \bf{48.82}  & 60.25 & \bf{46.25} \\
    \hline
    StarGAN & 59.68  & 71.07  & 51.36  & 78.87  & 99.03  & 176.18  & 70.40  & 142.35  & 70.14  & 206.21  \\
    \hline
    DNA-GAN & 79.27  & 76.89  & 77.04  & 72.35  & 126.33  & 127.66  & 75.02  & 75.96 &  121.04 & 118.67 \\
    \hline
        ELEGANT & 30.71  & \bf{31.12}  & 25.71  & 24.88  & \bf{37.51}  & \bf{49.13}  & 47.35  & 60.71  &  \bf{59.37} & 56.80 \\
    \tabucline[1.5pt]{-}
    \end{tabu}%
  \label{tab:FID}%
\end{table}%

\section{Conclusions}

We have established a novel model ELEGANT for transferring multiple face attributes. The model encodes different attributes into disentangled parts and generate images with novel attributes by exchanging certain parts of latent encodings. Under the observation that only local part of the image should be modified to transfer face attribute, we adopt the residual learning to facilitate training on high-resolution images. A U-Net structure design and multi-scale discriminators further improve the image quality. Experimental results on CelebA face database demonstrate that ELEGANT successfully overcomes three common limitations existing in most of other methods.

{\noindent\bf Acknowledgement.} This work was supported by High-performance Computing Platform of Peking University.

\bibliography{1583}

\begin{thebibliography}{10}
\providecommand{\url}[1]{\texttt{#1}}
\providecommand{\urlprefix}{URL }
\providecommand{\doi}[1]{https://doi.org/#1}

\bibitem{bengio2013better}
Bengio, Y., Mesnil, G., Dauphin, Y., Rifai, S.: Better mixing via deep
  representations. In: International Conference on Machine Learning. pp.
  552--560 (2013)

\bibitem{choi2018stargan}
Choi, Y., Choi, M., Kim, M., Ha, J.W., Kim, S., Choo, J.: Stargan: Unified
  generative adversarial networks for multi-domain image-to-image translation.
  IEEE Conference on Computer Vision and Pattern Recognition (CVPR)  (2018)

\bibitem{gardner2015deep}
Gardner, J.R., Upchurch, P., Kusner, M.J., Li, Y., Weinberger, K.Q., Bala, K.,
  Hopcroft, J.E.: Deep manifold traversal: Changing labels with convolutional
  features. arXiv preprint arXiv:1511.06421  (2015)

\bibitem{gatys2015a}
Gatys, L.A., Ecker, A.S., Bethge, M.: A neural algorithm of artistic style.
  Nature Communications  (2015)

\bibitem{goodfellow2014generative}
Goodfellow, I., Pouget-Abadie, J., Mirza, M., Xu, B., Warde-Farley, D., Ozair,
  S., Courville, A., Bengio, Y.: Generative adversarial nets. In: Advances in
  neural information processing systems. pp. 2672--2680 (2014)

\bibitem{gretton2012kernel}
Gretton, A., Borgwardt, K.M., Rasch, M.J., Sch{\"o}lkopf, B., Smola, A.: A
  kernel two-sample test. Journal of Machine Learning Research
  \textbf{13}(Mar),  723--773 (2012)

\bibitem{he2016dual}
He, D., Xia, Y., Qin, T., Wang, L., Yu, N., Liu, T., Ma, W.Y.: Dual learning
  for machine translation. In: Advances in Neural Information Processing
  Systems. pp. 820--828 (2016)

\bibitem{he2016deep}
He, K., Zhang, X., Ren, S., Sun, J.: Deep residual learning for image
  recognition. In: Proceedings of the IEEE conference on computer vision and
  pattern recognition. pp. 770--778 (2016)

\bibitem{heusel2017gans}
Heusel, M., Ramsauer, H., Unterthiner, T., Nessler, B., Hochreiter, S.: Gans
  trained by a two time-scale update rule converge to a local nash equilibrium.
  In: Advances in Neural Information Processing Systems. pp. 6629--6640 (2017)

\bibitem{isola2017image}
Isola, P., Zhu, J.Y., Zhou, T., Efros, A.A.: Image-to-image translation with
  conditional adversarial networks. arXiv preprint  (2017)

\bibitem{kim2017learning}
Kim, T., Cha, M., Kim, H., Lee, J.K., Kim, J.: Learning to discover
  cross-domain relations with generative adversarial networks. In: Precup, D.,
  Teh, Y.W. (eds.) Proceedings of the 34th International Conference on Machine
  Learning. Proceedings of Machine Learning Research, vol.~70, pp. 1857--1865.
  PMLR, International Convention Centre, Sydney, Australia (06--11 Aug 2017),
  \url{http://proceedings.mlr.press/v70/kim17a.html}

\bibitem{kingma2015adam}
Kingma, D.P., Ba, J.L.: Adam: A method for stochastic optimization.
  international conference on learning representations  (2015)

\bibitem{lample2017fader}
Lample, G., Zeghidour, N., Usunier, N., Bordes, A., DENOYER, L., et~al.: Fader
  networks: Manipulating images by sliding attributes. In: Advances in Neural
  Information Processing Systems. pp. 5963--5972 (2017)

\bibitem{li2016deep}
Li, M., Zuo, W., Zhang, D.: Deep identity-aware transfer of facial attributes.
  arXiv preprint arXiv:1610.05586  (2016)

\bibitem{liu2017unsupervised}
Liu, M.Y., Breuel, T., Kautz, J.: Unsupervised image-to-image translation
  networks. In: Advances in Neural Information Processing Systems. pp. 700--708
  (2017)

\bibitem{liu2015faceattributes}
Liu, Z., Luo, P., Wang, X., Tang, X.: Deep learning face attributes in the
  wild. In: Proceedings of International Conference on Computer Vision (ICCV)
  (2015)

\bibitem{lu2017conditional}
Lu, Y., Tai, Y.W., Tang, C.K.: Conditional cyclegan for attribute guided face
  image generation. arXiv preprint arXiv:1705.09966  (2017)

\bibitem{perarnau2016invertible}
Perarnau, G., van~de Weijer, J., Raducanu, B., {\'A}lvarez, J.M.: Invertible
  conditional gans for image editing. arXiv preprint arXiv:1611.06355  (2016)

\bibitem{reed2015deep}
Reed, S.E., Zhang, Y., Zhang, Y., Lee, H.: Deep visual analogy-making. In:
  Advances in neural information processing systems. pp. 1252--1260 (2015)

\bibitem{ronneberger2015u}
Ronneberger, O., Fischer, P., Brox, T.: U-net: Convolutional networks for
  biomedical image segmentation. In: International Conference on Medical image
  computing and computer-assisted intervention. pp. 234--241. Springer (2015)

\bibitem{shen2017learning}
Shen, W., Liu, R.: Learning residual images for face attribute manipulation.
  In: IEEE Conference on Computer Vision and Pattern Recognition (CVPR). pp.
  1225--1233. IEEE (2017)

\bibitem{simonyan2015very}
Simonyan, K., Zisserman, A.: Very deep convolutional networks for large-scale
  image recognition. International Conference on Learning Representations
  (2015)

\bibitem{taigman2016unsupervised}
Taigman, Y., Polyak, A., Wolf, L.: Unsupervised cross-domain image generation.
  arXiv preprint arXiv:1611.02200  (2016)

\bibitem{upchurch2016deep}
Upchurch, P., Gardner, J., Bala, K., Pless, R., Snavely, N., Weinberger, K.Q.:
  Deep feature interpolation for image content changes. arXiv preprint
  arXiv:1611.05507  (2016)

\bibitem{wang2017tag}
Wang, C., Wang, C., Xu, C., Tao, D.: Tag disentangled generative adversarial
  network for object image re-rendering. In: Proceedings of the Twenty-Sixth
  International Joint Conference on Artificial Intelligence, IJCAI. pp.
  2901--2907 (2017)

\bibitem{wang2018high}
Wang, T.C., Liu, M.Y., Zhu, J.Y., Tao, A., Kautz, J., Catanzaro, B.:
  High-resolution image synthesis and semantic manipulation with conditional
  gans. IEEE Conference on Computer Vision and Pattern Recognition (CVPR)
  (2018)

\bibitem{xiao2018dna}
Xiao, T., Hong, J., Ma, J.: Dna-gan: Learning disentangled representations from
  multi-attribute images. International Conference on Learning Representations,
  Workshop  (2018)

\bibitem{Yi_2017_ICCV}
Yi, Z., Zhang, H., Tan, P., Gong, M.: Dualgan: Unsupervised dual learning for
  image-to-image translation. In: The IEEE International Conference on Computer
  Vision (ICCV) (Oct 2017)

\bibitem{zhao2018modular}
Zhao, B., Chang, B., Jie, Z., Sigal, L.: Modular generative adversarial
  networks. arXiv preprint arXiv:1804.03343  (2018)

\bibitem{DBLP:conf/bmvc/ZhouXYFHH17}
Zhou, S., Xiao, T., Yang, Y., Feng, D., He, Q., He, W.: Genegan: Learning
  object transfiguration and attribute subspace from unpaired data. In:
  Proceedings of the British Machine Vision Conference (BMVC) (2017),
  \url{http://arxiv.org/abs/1705.04932}

\bibitem{zhu2016generative}
Zhu, J.Y., Kr{\"a}henb{\"u}hl, P., Shechtman, E., Efros, A.A.: Generative
  visual manipulation on the natural image manifold. In: European Conference on
  Computer Vision. pp. 597--613. Springer (2016)

\bibitem{zhu2017unpaired}
Zhu, J.Y., Park, T., Isola, P., Efros, A.A.: Unpaired image-to-image
  translation using cycle-consistent adversarial networks. Proceedings of
  International Conference on Computer Vision (ICCV)  (2017)

\bibitem{zhu2017toward}
Zhu, J.Y., Zhang, R., Pathak, D., Darrell, T., Efros, A.A., Wang, O.,
  Shechtman, E.: Toward multimodal image-to-image translation. In: Advances in
  Neural Information Processing Systems. pp. 465--476 (2017)

\end{thebibliography}
\bibliographystyle{splncs04}

\end{document}